\documentclass{article}

\usepackage[preprint]{neurips_2026}

\usepackage[utf8]{inputenc} 
\usepackage[T1]{fontenc}    
\usepackage{graphicx}       
\usepackage[
    colorlinks=true,
    linkcolor=blue!60!black,
    citecolor=blue!60!black,
    urlcolor=blue!70!black
]{hyperref}       
\usepackage{url}            
\usepackage{booktabs}       
\usepackage{array}          
\usepackage{amsmath}        
\usepackage{amsfonts}       
\usepackage{nicefrac}       
\usepackage{microtype}      
\usepackage{xcolor}         
\usepackage{placeins}       
\usepackage{caption}        
\usepackage{subcaption}     
\usepackage[most]{tcolorbox}
\usepackage{algorithm}      
\usepackage{algorithmic}    
\usepackage{multirow}
\usepackage[most]{tcolorbox}
\usepackage{xcolor}
\tcbuselibrary{listings,breakable}

\definecolor{sysbg}{HTML}{DCEFF7}
\definecolor{sysbd}{HTML}{9ECBE0}
\definecolor{usrbg}{HTML}{E5F7E5}
\definecolor{usrbd}{HTML}{9FD49F}

\newtcblisting{syspromptbox}{
  enhanced,
  breakable,
  listing only,
  colback=sysbg,
  colframe=sysbd,
  boxrule=0.5pt,
  arc=3pt,
  left=4pt, right=4pt, top=3pt, bottom=3pt,
  boxsep=1pt,
  width=\linewidth,
  listing options={
    basicstyle=\ttfamily\tiny,
    columns=fullflexible,
    keepspaces=true,
    breaklines=true,
    breakatwhitespace=false
  }
}

\newtcblisting{usrpromptbox}[1]{
  enhanced,
  breakable,
  listing only,
  colback=usrbg,
  colframe=usrbd,
  boxrule=0.5pt,
  arc=3pt,
  left=4pt, right=4pt, top=3pt, bottom=3pt,
  boxsep=1pt,
  width=\linewidth,
  title={\small\bfseries #1},
  coltitle=black,
  colbacktitle=usrbd!60!white,
  listing options={
    basicstyle=\ttfamily\tiny,
    columns=fullflexible,
    keepspaces=true,
    breaklines=true,
    breakatwhitespace=false
  }
}

\title{When Reasoning Narrows the Move: Diversity Collapse in LLM Game Play}

\author{%
  Junyi Sha\thanks{Equal Contribution, author ordered alphabetically.} \\
  Department of Civil and Environmental Engineering\\
  MIT \\
  Cambridge, MA 02139 \\
  \texttt{jsha@mit.edu} \\
  \And
  Renfei Tan\footnotemark[1]\\
  Institute for Data, Systems, and Society \\
  MIT \\
  Cambridge, MA 02139 \\
  \texttt{rftan@mit.edu} \\
  \And
  David Simchi-Levi \\
  Department of Civil and Environmental Engineering\\
  MIT \\
  Cambridge, MA 02139 \\
  \texttt{dslevi@mit.edu} \\
}

\begin{document}

\maketitle

\begin{abstract}
  Supervised fine-tuning (SFT) is widely used to adapt large language models to downstream tasks, but its effect on behavioral diversity in sequential decision-making remains under-explored.
  We study this question in a controlled suite of deterministic board games based on tic-tac-toe variants, where optimal actions are exactly computable and diversity can be measured directly.
  Across state-level evaluation, arena gameplay, and training trajectories, we find that reasoning-mode generation frequently suppresses action diversity without uniformly improving action accuracy. 
  Furthermore, standard SFT improves accuracy but often induces premature diversity collapse, which exceeds what is minimally required by the accuracy--diversity tradeoff.
  We then show that action augmentation, which trains on all optimal actions per state rather than a single demonstrated action, would partially mitigates this effect.
  Our results identify narrow-support imitation as a source of policy collapse in LLM decision-making and suggest that preserving action support during SFT is important for maintaining exploratory behavior.
\end{abstract}

\section{Introduction}

Recent advances in large language models (LLMs) have significantly expanded their capabilities in complex reasoning and task execution, motivating their use in decision-making settings. 
However, effectively aligning these models with task-specific objectives remains a key challenge.
A standard approach to improving downstream performance is Supervised Fine-Tuning (SFT).
SFT fine-tunes a pre-trained model on a set of high-quality prompt-response demonstrations, and throughout the process the model would imitate desired behavior.
SFT has proven highly effective, particularly for tasks requiring structured outputs and strict formats.

However, an important concern is that SFT may systematically suppress output diversity.
Because the training objective encourages the model to match the empirical distribution of the training data, it tends to favor reproducing patterns observed during training \citep{o2024attributing}.
In decision-making settings, this issue is particularly critical, as diversity underpins exploration, which is essential for discovering high-quality actions and adapting to previously unseen states \citep{hong2018diversity, li2024preserving}.
A loss of diversity may therefore limit performance and hinder subsequent post-training methods such as reinforcement learning \citep{kirk2023understanding, murthy2025one}.
Understanding how SFT impacts action diversity is therefore crucial for developing more effective training pipelines and improving LLM's decision-making abilities.

Games offer a clean and controlled environment for studying this question.
First of all, games have well-defined state, action, and reward spaces, which enables precise evaluation of policies.
In contrast, in real-world domains such as coding, theorem proving, or business planning, it is hard to define the performance and diversity of a policy.
In addition, one can separate performance (Did the model achieve a high score or favorable outcome?) from diversity (How broad was the set of actions the model actually played?), which is often confounded in messy field settings \citep{cipolina2025game, park2024llm}.
Finally, games allow large-scale experimentation involving repeated trajectories, making it feasible to systematically evaluate interventions such as prompting strategies or exploration incentives \citep{topsakal2024evaluating, xi2025agentgym, zhang2024simple}.

In this work, we focus on board games involving tic-tac-toe and its variants \citep{mishra2025ttt}, which are all two-player sequential games that are deterministic and with complete information.
Despite strong performance in many tasks, one may be surprised to find that LLM has performance only comparable to a one-step lookahead player even on the simplest and solved tic-tac-toe game \citep{duan2024gtbench, mishra2025ttt}.

Our central finding is that improved performance and reduced diversity are related, but not equivalent.
Reasoning-mode generation consistently compresses action diversity, often without a corresponding improvement in accuracy.
This suggests that reasoning does not merely make the model more optimal; it can also amplify weak action preferences and concentrate the policy around a single move.

We then study how this collapse evolves during fine-tuning.
Checkpoint-level trajectories show that diversity loss often occurs early, before the model reaches high accuracy.
Moreover, by comparing model trajectories to an iso-accuracy entropy ceiling, we find that much of the entropy loss exceeds what is minimally required by improved accuracy.
In other words, SFT does not only move the model along an accuracy-diversity tradeoff; it can push the policy below that frontier.

Finally, we test whether increasing action support in the SFT data can mitigate this effect.
Training on all optimal actions per state preserves substantially more diversity than training on a single demonstrated action.
However, action augmentation alone is not sufficient: the direct augmented model often remains less accurate.
The strongest configuration is the combination of action-augmented data and reasoning-mode generation, which achieves competitive accuracy while preserving non-trivial behavioral diversity.
Together, these results show that diversity collapse is a measurable and avoidable failure mode of narrow-support imitation in LLM decision-making.

\begin{figure}
    \centering
    \includegraphics[width=\linewidth]{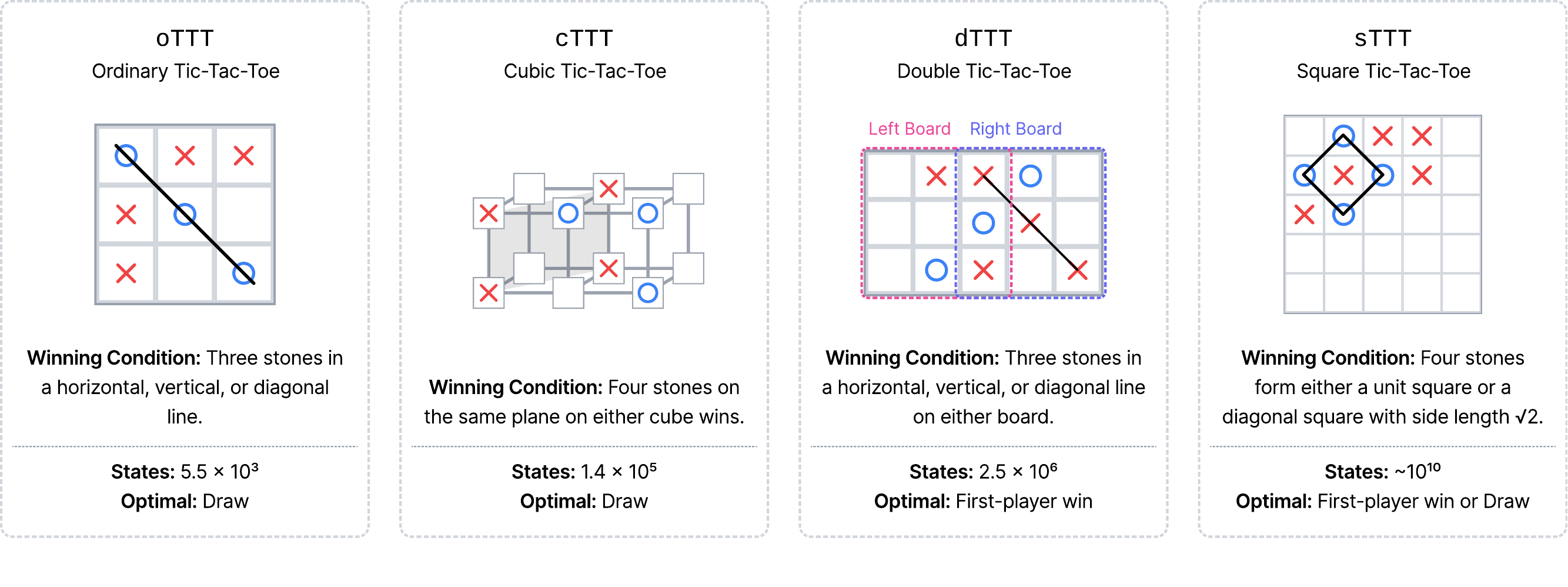}
    \caption{Overview of game environments.}
    \label{fig: games}
\end{figure}

\section{Related Work}

\paragraph{Large Language Model Game-playing.}
Many recent works have investigated LLM's performance in games. 
However, empirical studies show that LLMs remain surprisingly fragile in simple, fully-solved decision-making environments \citep{mishra2025ttt}. 
Across grid-based games \citep{topsakal2024evaluating, duan2024gtbench}, chess \citep{wen2025chessqa, kolasani2025llm} and imperfect-information card games \citep{su2025enhancing}, models often fail to reliably execute correct strategies. 
In grid-based games such as tic-tac-toe, \cite{duan2024gtbench} show that LLMs consistently underperform classical search-based agents, while \cite{topsakal2024evaluating} further find that failures are concentrated in immediate decision-making. 
Similar limitations extend beyond grid-based settings: \cite{van2025baba} show that fine-tuning improves rule analysis but not execution, while \cite{jiang2025marble} report near-zero accuracy on spatial planning tasks.
Taken together, this literature highlights that LLMs struggle to act reliably in sequential decision-making tasks. 
However, existing evaluations primarily focus on the quality of actions, leaving open the study on action diversity and its impact on performance.

\paragraph{Output Diversity in Supervised Fine-tuning.}
In supervised fine-tuning, models are trained to imitate curated examples \citep{ouyang2022training, bai2022training}. 
However, the same imitation objective that improve task performance can also narrow the model’s output distribution \citep{kirk2023understanding, murthy2025one}.
Recent work attributes this phenomenon to mode collapse during fine-tuning \citep{o2024attributing}, the mismatch between cross-entropy training and the existence of multiple valid outputs \citep{li2024preserving}, and the entropy-suppressing effects of structured formats \citep{yun2025price}.
To this end, multiple approaches have been proposed to preserve diversity including data diversification \citep{tirumala2023d4} and loss objective modification \citep{li2024preserving, chen2026sed, zhang2026good, xi2025agentgym}.
These findings motivate our study of SFT in games: unlike open-ended text generation, board games provide a finite and verifiable action space, allowing us to measure whether SFT collapses the policy toward a small set of demonstrated actions even when alternative legal or strategically valid actions exist.

\paragraph{Action Diversity in Offline Reinforcement Learning.}
Our setting is closely related to offline reinforcement learning (Offline RL), which studies the problem of learning decision-making policies from fixed datasets without further online interaction \citep{levine2020offline}.  
Classic work shows that offline RL suffer from covariate shift because small errors may move the learner into states that are poorly covered by the demonstration distribution \citep{ross2011reduction}, emphasizing that dataset coverage and distributional support are central to downstream policy quality \citep{fu2020d4rl}. 
Recent sequence-modeling approaches, including Decision Transformer, Trajectory Transformer, and RvS, further blur the boundary between supervised learning and offline RL by treating trajectory-conditioned decision-making as a supervised sequence-modeling problem \citep{chen2021decision, janner2021offline, emmons2021rvs}. 
Most relevant to our work, several recent papers study how diverse demonstrations or heterogeneous offline datasets can be leveraged to recover multiple high-quality behaviors rather than a single averaged policy \citep{yue2024leverage, mao2024stylized, petitbois2025offline}. 
This perspective suggests that action diversity determines the strategies represented in the training data and can affect whether subsequent optimization discovers better behavior.

\section{Game and Player Setup}
\label{sec:games-setup}
We design experiments to characterize how reasoning-mode output and supervised fine-tuning procedure affect an LLM policy's playing strength and behavioral diversity in board games. 
The setup uses four deterministic two-player turn-based games, four baselines policies per game, and six LLM players built from \texttt{Qwen3-8B} \citep{yang2025qwen3}.
Further implementation details for players, evaluation, SFT data generation and training, as well as metrics are collected in Appendix \ref{sec: appendix-implementation}.                                      

\paragraph{Preliminaries.}
We model each game deterministic as a Markov Decision Process (MDP)
$\langle \mathcal{S}, \mathcal{A}, P, r\rangle$, where $\mathcal{S}$ is the state space, $\mathcal{A}$ is the action space, $P$ fis the transition kernel together with initial state distribution, and $r$ is the reward function.
Two players alternately observe the current state $s_t$ and take actions $a_t$ according to their policies $\pi_A, \pi_B$.
After $a_t$ is taken, the game transitions to a new state $s_{t+1}$ according to $P$, and rewards are assigned through $r(s_t, a_t)$.
This process produces a trajectory $\tau = (s_1, a_1, \dotsb, s_T, a_T)$.
For all games considered in this work, rewards are only assigned at terminal states and players use discount factor $\gamma=1$.
Rewards of $1$, $1/2$, and $0$ correspond to win, draw, and loss respectively.

\paragraph{Environments and Players.}
We consider standard Tic-Tac-Toe (\texttt{ottt}) and three variants proposed by \citep{mishra2025ttt}: \texttt{cttt}, \texttt{dttt}, and \texttt{sttt}.
An overview of the games is shown in Figure~\ref{fig: games}.
We select these games for three reasons.
First, the rules, states, and actions are simple and can be represented naturally in text prompts.
Second, despite their simplicity, the games contain nontrivial tactical patterns and remain challenging for LLMs \citep{mishra2025ttt}.
Third, all games can be solved optimally through recursive search, which provides a clear performance reference.

For each game, we construct four baseline players and six LLM players.
The baseline players span different levels of playing strength and action diversity, ranging from random play to minimax-optimal play.
Details of the baseline policies are provided in Appendix \ref{subsec: app-arena-players}.
For LLM players, we focus on \texttt{Qwen3-8B} and its fine-tuned variants.
We additionally evaluate other open-weight models of similar scale in Appendix \ref{subsec: app-other-baseline-LLMs}.
We consider two output modes: direct generation and reasoning generation.
For each mode, we evaluate the base model together with two fine-tuned variants, resulting in six LLM players in total.
All players share the same prompting framework.
A fixed system prompt specifies the required output format.
The user prompt contains the game rules, current board state, and legal actions.
Only the current state is provided.
Past trajectories are omitted so that the model acts purely as a Markovian policy $\pi(a_t \mid s_t)$ conditioned only on the current state.
In Appendix \ref{sec: app-prompt-engineering}, we experiment with \texttt{GEPA} prompt optimization \citep{agrawal2025gepa} and show that prompt optimization alone is insufficient for strong gameplay performance.

\paragraph{Supervised Fine-tuning.}
Our experiments focus on how SFT affects action diversity.
For each game, we construct four SFT datasets that share the same state pool and the same total number of training rows.
The state pool is generated through Monte-Carlo random self-play.
Each sampled state is solved using the strongest baseline policy.
We then reduce the collected states to a fixed training budget using visit-count selection.

The datasets vary along two dimensions:
(i) action support, namely whether each state is paired with a single optimal action or all optimal actions, and
(ii) output format, namely whether the response contains only the final action (direct) or includes an explicit reasoning trace (reasoning).
Combining these two dimensions yields four dataset variants: \texttt{plain-direct}, \texttt{aug-direct}, \texttt{plain-reason}, and \texttt{aug-reason}.
Here, \texttt{plain} denotes datasets containing one demonstrated optimal action per state, while \texttt{aug} includes all optimal actions returned by the solver.
Similarly, \texttt{direct} contains only the final action output, whereas \texttt{reason} additionally includes a reasoning trace before the action.
This design isolates the effects of action support and reasoning format while controlling for state coverage and total training tokens.
For reasoning datasets, the reasoning trace evaluates every legal action and annotates its tactical consequence under the solver, such as immediate wins, forced blocks, fork creation, or balanced continuations.
The trace then concludes with the set of optimal actions.

We fine-tune \texttt{Qwen3-8B} using LoRA and save one adapter after each epoch.
The final-epoch adapter is used as the deployed SFT player.
We additionally retain three intermediate checkpoints from each run for the diversity dynamics analysis in \ref{sec:trajectory}.
Training hyperparameters are listed in Appendix \ref{app:sft-hyperparameters}.

\section{Main Experiments: Policy Evaluation and Game-Playing}

\subsection{Policy Analysis}
\label{subsec: policy-analysis}

To study what each LLM player has learned at the state level, we evaluate every player on a curated state pool of 200 board states per game, consisting of 100 training states and 100 held-out (``new'') states.
For each state, we sample $N=8$ independent action predictions. 
Trained states are sampled from the SFT state pool used for fine-tuning.
New states are generated through random self-play and de-duplicated against the training set.
Within each pool, we balance five tactical categories with increasing reasoning depth: \texttt{win}, \texttt{block}, \texttt{fork}, \texttt{fork-deny}, and \texttt{positional}.
We additionally balance \texttt{x}-to-move and \texttt{o}-to-move positions.
We exclude states in which every legal action is optimal, such as fully symmetric forced-draw positions, since these states trivialize accuracy evaluation.

\begin{figure*}[t]
  \centering
  \includegraphics[width=\linewidth]{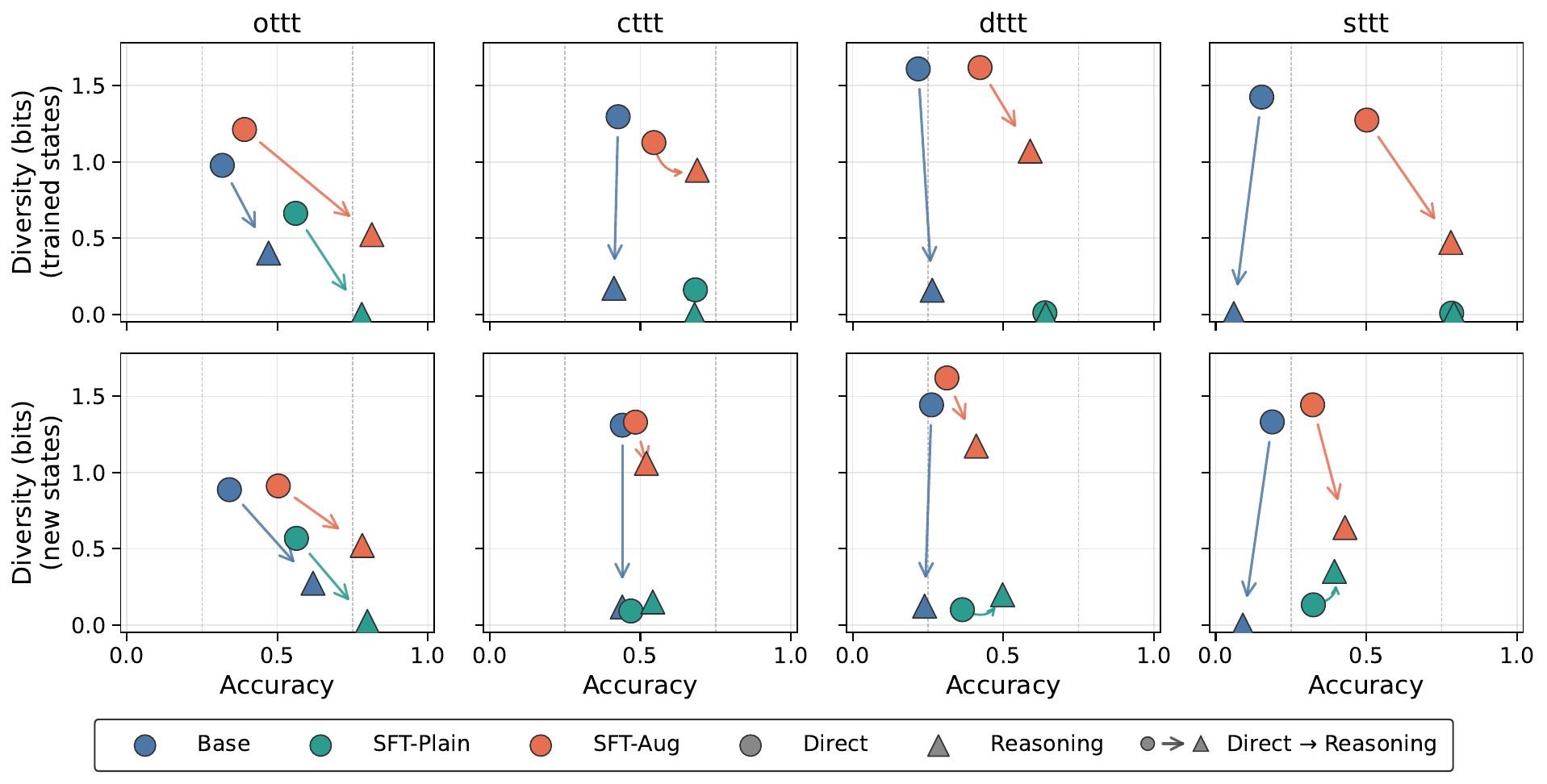}
  \caption{
    State-level accuracy-diversity profiles per game on \emph{trained} states (top row) and \emph{new} states (bottom row). 
    Each panel plots the six model configurations, with arrows connecting direct to reasoning mode within the same variant. 
    Rightward movement indicates higher accuracy, while downward movement indicates lower empirical action entropy.}

  \label{fig:part2-acc-div}
\end{figure*}

We evaluate both accuracy and \textit{action diversity}.
For a policy $\pi$, we define action diversity as the expected action entropy
$\mathbb{E}_{S \sim \mu}[H(\pi(\cdot \mid S))]$
under either the training-state distribution $\mu_{\text{train}}$ or the new-state distribution $\mu_{\text{new}}$.
The results are shown in Figure \ref{fig:part2-acc-div}.

\paragraph{Reasoning suppresses diversity without uniformly improving accuracy.}
Switching from direct to reasoning mode is associated with a substantial reduction in action diversity.
Across the trained and new state pools, reasoning exhibits lower entropy than direct mode in 21 out of 24 scenarios; in the remaining three cases, both modes already produce nearly deterministic outputs.
Most direct-to-reasoning transitions are dominated by downward movement rather than rightward movement.
This effect is especially pronounced in the base model, on which we observe the most significant reduction in diversity.
Nevertheless, this reduction is not consistently accompanied by higher accuracy.
Substantial accuracy improvements are observed only in game \texttt{ottt}, while the other three variants show little or even negative effect from reasoning.
Thus, the lower diversity of reasoning-mode players cannot be explained purely as a consequence of more optimal (and therefore more convergent) play.

One possible explanation is that the model already contains weak preferences over actions before reasoning begins.
During reasoning, intermediate evaluations and justifications repeatedly reinforce these preferences, even when multiple actions remain strategically equivalent.
As the reasoning trajectory progresses, small initial asymmetries may therefore become amplified, eventually concentrating probability mass onto a single action.

\paragraph{Plain SFT improves accuracy but induces near mode-collapse behavior.}
Standard supervised fine-tuning substantially improves state-level accuracy relative to the Base model, especially on trained states.
However, this improvement is accompanied by a sharp reduction in action diversity.
The effect is particularly strong when combined with reasoning mode.
Across games, \texttt{plain-reason} achieves strong accuracy on trained states but exhibits near-zero entropy, indicating an almost deterministic policy despite the existence of multiple strategically equivalent actions.
Even \texttt{plain-direct} produces substantially lower diversity than the Base direct model.

These results suggest that plain SFT drives the model toward a near mode-collapse regime, where the policy memorizes and repeatedly selects a single preferred action for each state.
Reasoning further amplifies this concentration effect, causing the policy distribution to collapse even when the accuracy improvement is marginal.

\paragraph{Reasoning combined with action-augmented SFT preserves non-trivial diversity.}
Action-augmented SFT partially mitigates this diversity collapse by exposing the model to multiple optimal actions during training.
Compared with \texttt{plain-reason}, \texttt{aug-reason} achieves similar accuracy on trained states while retaining substantially higher entropy across all games.
The same pattern persists on new states, where \texttt{aug-reason} remains competitive in accuracy while preserving significantly more diversity than \texttt{plain-reason}.

Notably, reasoning combined with action-augmented training is the only setting that consistently maintains non-trivial diversity at relatively high accuracy.
This suggests that increasing the demonstrated action support during SFT can partially decouple performance improvement from policy collapse.

\subsection{Arena Game-play}
\label{subsec: arena-gameplay}

State-level evaluation provides a controlled view of a model's local decision-making ability, but it does not fully characterize actual game-play.
During a game, the trajectory depends jointly on both players, and many states may never be visited.
To evaluate realized gameplay behavior, we let the four baseline players and six LLM players compete in an arena setting.
For each pair of baseline players, we run 1000 games with one player assigned to \texttt{x} and the other to \texttt{o}, followed by another 1000 games with roles reversed.
For each pair consisting of one LLM player and one baseline player, we run $100+100$ games under the two role assignments.
The six LLM players do not play against each other, ensuring that all LLM evaluations use the same controlled opponent pool.

\begin{figure}[h]
    \centering
    \includegraphics[width=0.9\linewidth]{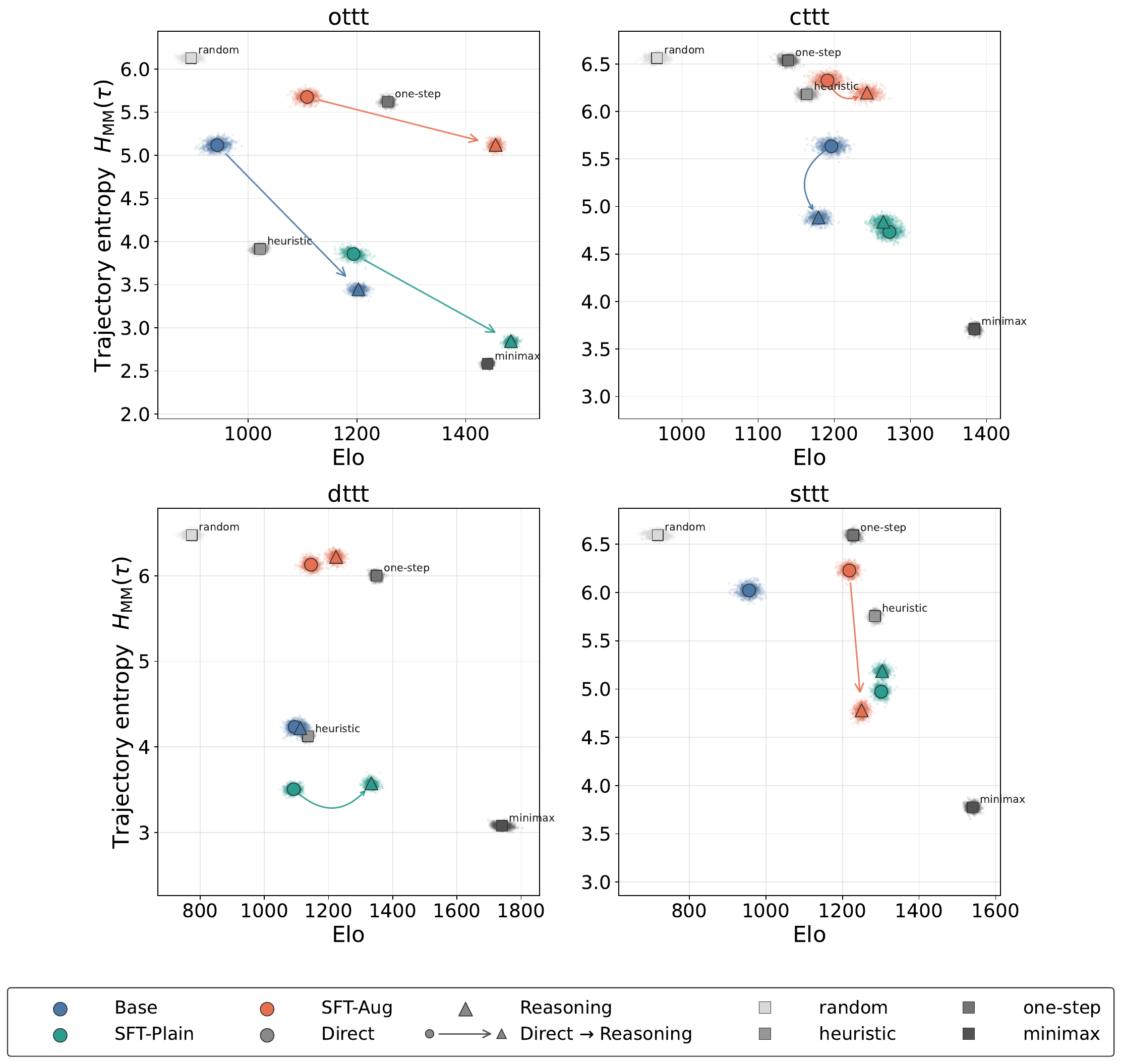}
    \caption{Arena-level Elo-diversity profiles per game. 
    For each game, we visualize the joint tradeoff between playing strength and behavioral diversity. 
    The six LLM configurations are plotted alongside four reference baselines: \texttt{random}, \texttt{heuristic}, \texttt{one-step}, and \texttt{minimax}. 
    Arrows connect each direct-mode model to its reasoning-mode counterpart within the same variant. 
    Rightward movement corresponds to higher Elo, whereas downward movement corresponds to lower trajectory entropy. 
    Transparent point clouds show the empirical spread of $B=1000$ stratified bootstrap replicates around each estimate.}
    \label{fig:arena}
\end{figure}

\paragraph{Metrics.}
We evaluate playing strength using the Elo rating system \citep{elo1978rating}, which is widely used in competitive games such as Chess and Go.
Each player is assigned a scalar rating, and rating differences determine expected game outcomes.
For two players with ratings $R_A$ and $R_B$, the expected reward of player $A$ is $\mathbb E[r_A] = 1/(1 + \exp(-\ln 10(R_A - R_B)/\tau))$.

Beyond Elo, we also measure behavioral diversity through trajectory entropy.
Given two policies $\pi_A$ and $\pi_B$, their interaction induces a trajectory distribution $p_{\pi_A,\pi_B}(\tau)$ over complete game trajectories.
We define trajectory entropy as $H_{\pi_A, \pi_B}(\tau) = -\sum_{\tau} p_{\pi_A, \pi_B}(\tau)\log p_{\pi_A, \pi_B}(\tau)$.
Against an opponent pool ${\pi_1,\dots,\pi_K}$ with mixture weights $p_k$, diversity is measured by the conditional entropy $H(\tau\mid K)$ where $K\sim p_k$.
We note that the trajectory entropy decomposed into accumulated action entropy of $\pi_A$ and $\pi_B$ under an induced state distribution $\mu(s)$:
\begin{equation}
    H(\tau) = \sum_{t=1}^T H(a_t \mid s_{t-1}) = \sum_{i=1}^{\lfloor T/2\rfloor} \sum_s \mu_{2i}(s) H(\pi_A(\cdot \mid s)) + \mu_{2i+1}(s) H(\pi_B(\cdot \mid s)).
\end{equation}
From this decomposition, one can see that trajectory entropy reflects both the stochasticity of player policies and the state distribution induced by gameplay.
It is therefore not purely a property of a single policy.
Nevertheless, it captures realized behavioral diversity against a fixed opponent pool, which we view as a practically meaningful measure of gameplay diversity.

Performance and diversity are also not independent.
Policies with stronger play often concentrate probability mass on a narrower set of strategically strong trajectories, naturally reducing entropy.
However, our experiments reveal many LLM players with very low diversity despite relatively weak performance.
This suggests that premature diversity collapse can emerge even far from optimal play, and therefore cannot be explained solely by a performance-diversity tradeoff.

\paragraph{Bootstrap Uncertainty.}
To quantify uncertainty in arena evaluation, we use a stratified nonparametric bootstrap.
For each condition, games are partitioned by opponent and playing role.
Within each stratum, games are resampled with replacement at the original sample size.
For each bootstrap replicate $b=1,\dots,B$, we recompute the statistic of interest $\hat{\theta}_c^{(b)}$, either Elo or trajectory entropy.
For comparisons between conditions $c$ and $c'$, we compute the bootstrap contrast
$\hat{\Delta}^{(b)}=\hat{\theta}c^{(b)}-\hat{\theta}{c'}^{(b)}$.


\paragraph{Results and Conclusions.}
Figure~\ref{fig:arena} shows that the main patterns identified in Section~\ref{subsec: policy-analysis} persist under realized gameplay trajectories.
Across games, transitions from direct to reasoning mode are typically dominated by reductions in trajectory entropy rather than improvements in Elo.
In several cases, reasoning substantially lowers diversity while producing only marginal gains in playing strength.

Similarly, plain SFT improves Elo relative to the Base model but often drives the policy toward low-diversity regimes.
This effect is especially pronounced for reasoning-mode players, which frequently exhibit sharp entropy collapse despite remaining far from minimax-level performance.
These observations further support the hypothesis that standard SFT induces premature policy concentration rather than merely reflecting convergence toward optimal play.

Compared with \texttt{plain} SFT variants, \texttt{aug} SFT consistently preserves higher trajectory entropy while maintaining competitive Elo.
In particular, \texttt{aug-reason} is the only configuration that repeatedly achieves relatively strong gameplay performance together with non-trivial behavioral diversity.
This suggests that exposing the model to multiple optimal actions during fine-tuning can partially counteract the entropy collapse induced by standard imitation learning and reasoning.

\section{Diversity Dynamics During Fine-tuning}
\label{sec:trajectory}

The previous section evaluates only the final deployed SFT adapter.
Two questions remain unanswered:
\emph{when} during fine-tuning does the diversity collapse identified in Figures~\ref{fig:part2-acc-div} and \ref{fig:arena} emerge, and
\emph{how much} of that collapse is mechanically induced by improved accuracy as opposed to representing additional policy concentration beyond the necessary accuracy-diversity tradeoff.
To study these questions, we re-run the policy evaluation from Section~4.1 on intermediate checkpoints collected throughout fine-tuning.

\begin{figure*}[tb]
\centering

\begin{subfigure}{\linewidth}
\includegraphics[width=\linewidth]{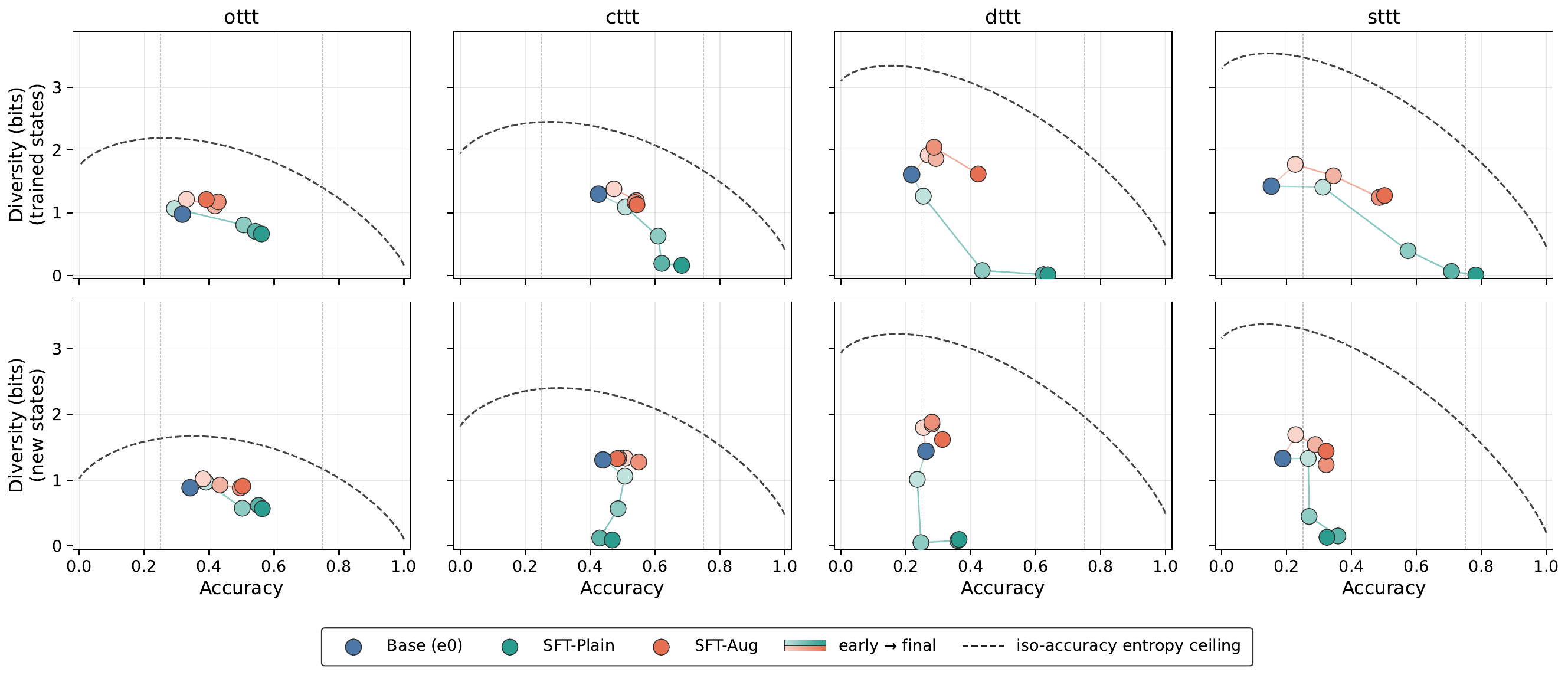}
\caption{Direct mode.}
\label{fig:part3-direct}
\end{subfigure}

\begin{subfigure}{\linewidth}
\includegraphics[width=\linewidth]{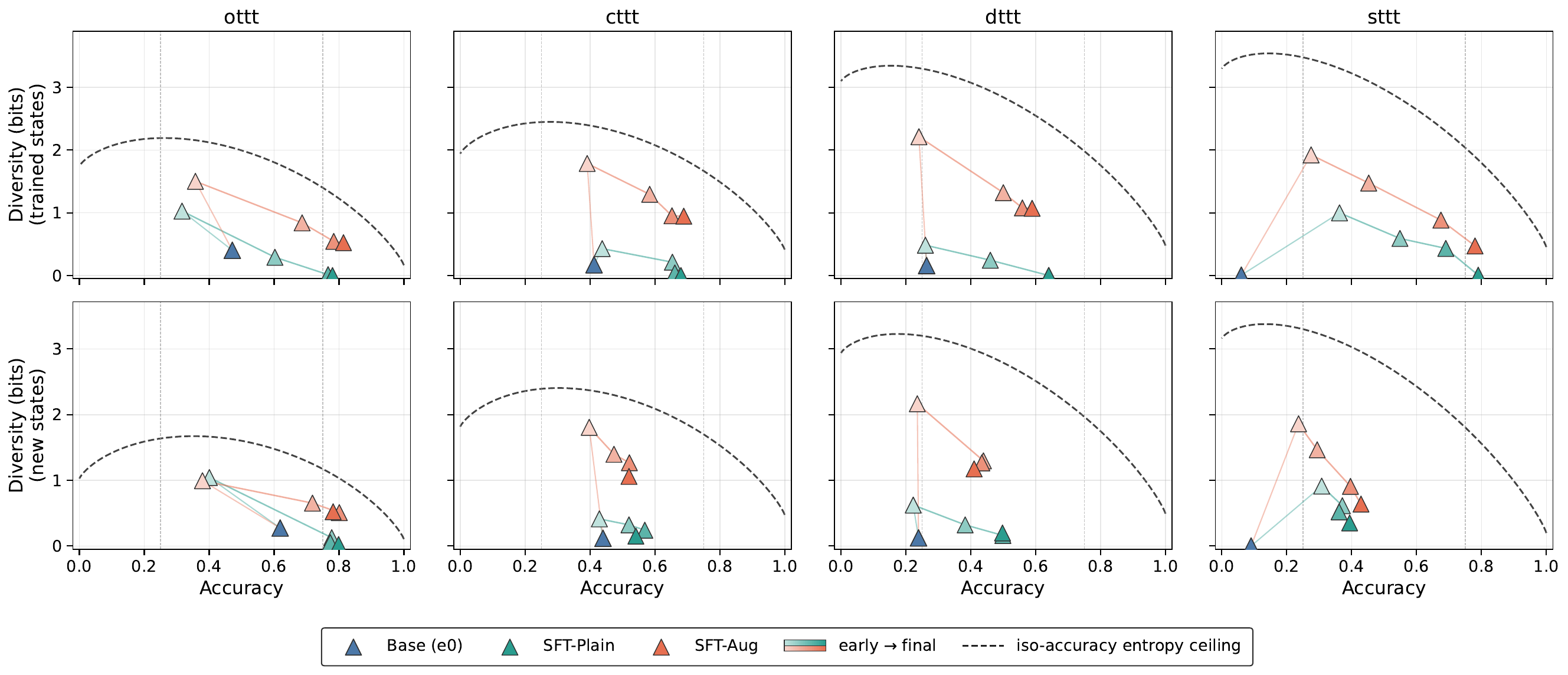}
\caption{Reasoning mode.}
\label{fig:part3-reasoning}
\end{subfigure}

\caption{
Per-game accuracy-diversity trajectories across SFT epochs on \emph{trained} states (top row of each subfigure) and \emph{new} states (bottom row).
Lighter shades indicate earlier epochs, while full saturation corresponds to the final adapter.
The Base model (blue) is plotted at $\text{epoch}=0$ and serves as the common starting point for both SFT variants.
The dashed black curve denotes the iso-accuracy entropy ceiling of~\eqref{eq:per-state-ceiling}, representing the maximum mean action entropy achievable at a given mean accuracy on the corresponding state pool.
}
\label{fig:part3-trajectory}

\end{figure*}

\paragraph{Experimental setup.}
For every $(\text{game}, \text{variant}, \text{mode})$ configuration, we evaluate three intermediate LoRA adapters in addition to the final checkpoint analyzed in Section~4.1.
For \texttt{ottt}, \texttt{cttt}, and \texttt{dttt}, we evaluate epochs $\{1,6,12,18\}$.
For \texttt{sttt}, where training is shorter, we evaluate epochs $\{1,4,8,12\}$.
The pre-fine-tuning Qwen3-8B base model provides the common $\text{epoch}=0$ initialization point for both Direct and Reasoning modes.
All other evaluation settings remain identical to Section~4.1, including the same curated 200-state evaluation pool, $N=8$ action samples per state, and identical decoding hyperparameters.
This produces a five-point trajectory in the accuracy-diversity plane for each SFT variant and output mode within every game.

\paragraph{Iso-accuracy entropy ceiling.}
To distinguish accuracy-induced concentration from additional diversity collapse, we compare each model against an analytic upper envelope that we call the \emph{iso-accuracy entropy ceiling}.
The ceiling answers the following question:
\emph{among all policies with a fixed accuracy level, what is the maximum achievable action entropy?}
Consider a state $s$ with $k_s$ optimal actions among $n_s$ legal actions.
Let $p_s$ denote the policy's probability mass assigned to optimal actions, i.e., its per-state accuracy.
For fixed $p_s$, entropy is maximized when probability mass is distributed uniformly across the optimal-action set and uniformly across the sub-optimal set.
This yields the per-state entropy ceiling

\begin{equation}
H_s^{\max}(p_s)
=
h_2(p_s)
+
p_s\log_2 k_s
+
(1-p_s)\log_2(n_s-k_s),
\label{eq:per-state-ceiling}
\end{equation}

where $h_2(\cdot)$ denotes the binary entropy function.
Averaging this quantity across the evaluation state pool produces the dashed black curve in Figure~\ref{fig:part3-trajectory}.
No policy with the same mean accuracy can achieve higher mean entropy on that state pool.
Therefore, the gap between a model trajectory and the ceiling measures diversity loss beyond what is mechanically required by improved accuracy.

\paragraph{Diversity collapse emerges early during fine-tuning and exceeds the accuracy-induced tradeoff.}
Across games and evaluation pools, most trajectories move rapidly downward in the accuracy--diversity plane during the first few epochs of fine-tuning, indicating that diversity collapses substantially faster than accuracy improves.
This effect is especially pronounced for \texttt{plain} SFT variants, which often approach near-deterministic behavior long before reaching high accuracy.
In contrast, \texttt{aug} variants consistently remain closer to the iso-accuracy entropy ceiling throughout training, preserving substantially more diversity at comparable accuracy levels.
The gap between observed trajectories and the ceiling further shows that much of the entropy reduction cannot be explained purely by improved accuracy or convergence toward optimal play.
Instead, standard SFT appears to induce additional policy concentration beyond what is mechanically required by the accuracy-diversity tradeoff, whereas action-augmented training partially mitigates this effect.

\section{Conclusion}

We study how supervised fine-tuning shapes action diversity in LLM decision-making using a controlled suite of board games.
Across state-level policy evaluation, arena gameplay, and checkpoint-level training trajectories, we find that standard SFT improves accuracy but often drives the policy toward premature diversity collapse.
This collapse is especially strong under reasoning-mode generation and is not fully explained by the accuracy-diversity tradeoff: models often lose more entropy than is necessary to achieve their observed accuracy.
Action-augmented SFT, which trains on multiple optimal actions per state, partially mitigates this collapse and preserves non-trivial diversity at comparable performance.
These results suggest that narrow-support imitation can make LLM policies overly concentrated even in simple sequential decision problems. Our study is limited to small deterministic games and a single main model family. 
In real-world applications such as recommendation, planning, and autonomous agents, similar diversity collapse may lead to overly concentrated behaviors, reduced exploration, and loss of valid alternatives. 
This could limit robustness and adaptability, highlighting the importance of training objectives that preserve action diversity, and motivating further research on diversity-aware training and evaluation in more complex decision-making settings.

\bibliographystyle{plainnat}
\bibliography{reference}

\newpage
\appendix
{\Large\bfseries Appendix\par}
\vspace{0.3em}
\addcontentsline{toc}{section}{Appendix}

\section{Implementation Details }
\label{sec: appendix-implementation}

\subsection{Arena Players}
\label{subsec: app-arena-players}

Table~\ref{tab:games-players} summarizes the players introduced in our arena: four deterministic policies and three open-source language models. We select the deterministic policies to provide interpretable anchors that span both skill and behavioral diversity, yielding broad coverage of the Elo-versus-diversity 2D space. We intentionally use open-source (open-weight) LLMs rather than closed proprietary APIs so that all models can be fine-tuned under a consistent training setup, enabling controlled analysis of how fine-tuning changes both playing strength and decision diversity while preserving reproducibility.

\FloatBarrier
\begin{table}[H]
	\centering
	\footnotesize
	\setlength{\tabcolsep}{6pt}
	\begin{tabular}{@{}>{\raggedright\arraybackslash}p{0.26\textwidth} >{\raggedright\arraybackslash}p{0.68\textwidth}@{}}
		\toprule
		\textbf{Player} & \textbf{Description} \\
		\midrule
		\texttt{random} &
		Uniform random choice among legal cells at each decision. Anchors low skill and a high-entropy player for calibrating diversity metrics. \\
		\texttt{heuristic} &
        Scores each legal cell by its participation count across all winning lines/planes (game-specific); samples uniformly among the highest-scoring cells \\
		\texttt{one\_step} &
		Takes immediate win if available; otherwise blocks opponent immediate win;
        otherwise falls back to a default legal-action policy \\
		\texttt{minimax} &
		Exact game-theoretic search for ottt/cttt/dttt (value-optimal; depth-to-mate
        tie-breaking); depth-limited negamax with alpha-beta pruning, transposition
        table, center-first move ordering, and a positional heuristic at the depth
        cutoff ($d{=}4$) for sttt  \\
		\midrule
		\texttt{Qwen3-8B} &
		Open-weight Alibaba Qwen~3 checkpoint (8B parameters), decoder-only. \\
		\texttt{Gemma-3-12B} &
		Open-weight Google Gemma~3 checkpoint (12B parameters), decoder-only. \\
		\texttt{Llama-3.1-8B} &
		Open-weight Meta Llama~3.1 checkpoint (8B parameters), decoder-only. \\
		\bottomrule
	\end{tabular}
    \vspace{3pt}
	\caption{Players in the arena experiments: four deterministic policies (top) and three open LLM agents (bottom).}
	\label{tab:games-players}
\end{table}




\newtcblisting{promptbox}[2][]{
  enhanced,
  breakable,
  listing only,
  colback=#1,
  colframe=#2,
  boxrule=0.5pt,
  arc=3pt,
  left=4pt,
  right=4pt,
  top=3pt,
  bottom=3pt,
  boxsep=1pt,
  width=\linewidth,
  listing options={
    basicstyle=\ttfamily\tiny,
    columns=fullflexible,
    keepspaces=true,
    breaklines=true,
    breakatwhitespace=false
  }
}

\subsection{Prompts for Different Games}
\label{app:qwen3-prompts}

For Qwen3, the system prompt and per-turn user prompt were identical in the thinking and non-thinking settings; the only difference was whether the chat template was rendered with thinking enabled. For each game, the Qwen3 prompt consisted of two parts: a system message and a per-turn user message. The system message was shared across all games, while the per-turn user message specified the rules and game state for the particular game being played. The exact prompts used are listed below.

\subsubsection*{Shared System Message}

\begin{syspromptbox}
You are a strategic game-playing assistant.
Follow the user's game rules exactly.
You may reason step-by-step before answering. After your reasoning, output exactly one action block of the form:
<action>{"move": <value>}</action>
The JSON inside <action> must be valid, and <value> must be one of the legal actions.
\end{syspromptbox}

\subsection*{Per-turn User Messages}

\begin{usrpromptbox}{oTTT}
You are playing OrdinaryTicTacToe.
Rules:
Standard 3x3 Tic-Tac-Toe with letter labels. Players X and O alternate placing their mark on any empty cell. A player wins by occupying any full row, full column, or full diagonal (three in a line). If all 9 cells fill without any player forming a line, the game is a draw. X moves first.
Cells are labeled A-I as:
A B C
D E F
G H I
Winning lines (8 total): rows ABC, DEF, GHI; columns ADG, BEH, CFI; diagonals AEI, CEG.
Example: if X holds {A, B} and plays C, X completes row ABC and wins.
You are: <role>
Board:
<row1>
---------
<row2>
---------
<row3>
Cell labels:
A B C
D E F
G H I
Legal actions: <legal>
After any reasoning, emit exactly one action block of the form:
<action>{"move": <action_label>}</action>
The JSON key must be "move" and the value must be one of the legal actions.
\end{usrpromptbox}

\begin{usrpromptbox}{dTTT}
You are playing DoubleTicTacToe.
Rules:
Double Tic-Tac-Toe has 15 cells: a 3x3 grid (A-I) plus a 3x2 extension to the right (J-O), laid out as:
A B | C | J K
D E | F | L M
G H | I | N O
Players X and O alternate placing their mark on any empty cell. X moves first. A player wins by completing any one of the 15 winning lines (three cells in a line). If all 15 cells fill without any player completing a line, the game is a draw.
Winning lines: ABC, DEF, GHI, ADG, BEH, CFI, AEI, CEG, CJK, FLM, INO, JLN, KMO, CLO, KLI.
Example: if X holds {A, B} and plays C, X completes line ABC and wins.
You are: <role>
Board:
<a> <b> | <c> | <j> <k>
<d> <e> | <f> | <l> <m>
<g> <h> | <i> | <n> <o>
Labels:
A B | C | J K
D E | F | L M
G H | I | N O
Legal actions: <legal>
After any reasoning, emit exactly one action block of the form:
<action>{"move": <action_label>}</action>
The JSON key must be "move" and the value must be one of the legal actions.
\end{usrpromptbox}

\begin{usrpromptbox}{cTTT}
You are playing CubeTicTacToe.
Rules:
Cube Tic-Tac-Toe is played on the 12 corners of two unit cubes that share one square face, forming a 2x1x1 cuboid. The shared-face corners are B, C, G, F; the left cube also has A, D, H, E, and the right cube also has I, J, L, K. Players X and O alternate placing their mark on any empty corner. X moves first. A player wins by fully occupying all four corners of any one of the 23 winning rectangles in 3D space. If all 12 corners fill without any rectangle being fully owned, the game is a draw.
Winning rectangles (23 total):
Face (11): ABCD, EFGH, ABFE, BCGF, CDHG, ADHE, BIJC, FKLG, BIKF, IJLK, JCGL
Diagonal (12): ACGE, BDHF, ABGH, DCFE, ADGF, BCHE, BJLF, ICGK, BILG, CJKF, BCLK, IJFG
Example: if X holds {A, C, G} and plays E, X completes diagonal rectangle ACGE and wins.
You are: <role>
Occupied by X: <x-cells>
Occupied by O: <o-cells>
Winning rectangles (23 total):
Face (11): ABCD, EFGH, ABFE, BCGF, CDHG, ADHE, BIJC, FKLG, BIKF, IJLK, JCGL
Diagonal (12): ACGE, BDHF, ABGH, DCFE, ADGF, BCHE, BJLF, ICGK, BILG, CJKF, BCLK, IJFG
Board labels: A, B, C, D, E, F, G, H, I, J, K, L
Legal actions: <legal>
After any reasoning, emit exactly one action block of the form:
<action>{"move": <action_label>}</action>
The JSON key must be "move" and the value must be one of the legal actions.
\end{usrpromptbox}

\begin{usrpromptbox}{sTTT}
You are playing SquaresTicTacToe.
Rules:
Squares Tic-Tac-Toe is played on a 5x5 grid (cells labeled A-Y, row-major). Players X and O alternate placing their mark on any empty cell. X moves first. A player wins by occupying 4 cells that form a square. A square may be either:
  (a) an axis-aligned 2x2 square (four corners of a unit square), or
  (b) a tilted (diamond) square of any size that fits inside the 5x5 board.
Label layout (row-major):
A B C D E
F G H I J
K L M N O
P Q R S T
U V W X Y
If all 25 cells fill without any player forming a square, the game is a draw.
Example (2x2): {A, B, F, G} is a square.
Example (diamond): {C, G, I, M} is a square (C top, G left, I right, M bottom).
You are: <role>
Board:
<row1>
<row2>
<row3>
<row4>
<row5>
Labels:
A B C D E
F G H I J
K L M N O
P Q R S T
U V W X Y
Legal actions: <legal>
After any reasoning, emit exactly one action block of the form:
<action>{"move": <action_label>}</action>
The JSON key must be "move" and the value must be one of the legal actions.
\end{usrpromptbox}

\subsection{SFT Dataset Construction: Base, Augmented, Thinking, and Thinking-Augmented}
\label{app:sft-dataset-construction}

For each game, we generate four supervised fine-tuning datasets from the same selected state pool, with identical row count and aligned state order across variants. The only differences are the target action choice and whether synthetic reasoning is included.

\paragraph{Step 1: Monte Carlo state pool.}
We run random self-play rollouts and collect non-terminal states until a game-specific unique-state target is reached (or a maximum rollout limit is hit).  
For each visited state, we store the board snapshot, current player, and visit count.

\paragraph{Step 2: Solver labels.}
Each pooled state is solved with the game solver to obtain:
\begin{itemize}
\item the optimal action set \(A_s^\star\),
\item a canonical best action (the first optimal action),
\item solver-side metadata (e.g., optimal-play outcome and its numeric encoding).
\end{itemize}

\paragraph{Step 3: Popularity-based selection under a row budget.}
For each game and split, we set a \emph{row budget} \(B\), i.e., the target number of training examples to generate for each dataset variant.  
States are sorted by Monte Carlo visit count (descending), then selected in that order.  
For a selected state \(s\), let \(K_s = |A_s^\star|\) be the number of optimal actions. We allocate \(K_s\) rows to that state.  
Selection continues until the cumulative row count reaches \(B\). If adding the final state would exceed \(B\), we truncate that state's allocation to the remaining rows. Therefore,
\[
\sum_{s \in \mathcal{S}_{\text{selected}}} K_s = B.
\]

\paragraph{Step 4: Construct four supervision variants.}
From the same selected state set, we construct four aligned supervision variants by crossing two factors:

\begin{itemize}
\item \textbf{Action supervision type}: \emph{single-target} vs. \emph{multi-target}.  
Single-target uses one canonical optimal move per state, while multi-target exposes all optimal moves as valid supervision.
\item \textbf{Reasoning content}: \emph{without reasoning} vs. \emph{with reasoning}.  
The reasoning variants prepend a brief explanatory rationale before the final action.
\end{itemize}

This yields four datasets: (i) single-target, (ii) multi-target (action-augmented), (iii) single-target with reasoning, and (iv) multi-target with reasoning.  
All four are derived from the same underlying state distribution so that comparisons isolate the effects of action augmentation and reasoning augmentation.

\paragraph{Prompt and row format (all variants).}
All variants share the same prompt structure and row schema: each example uses a common system-message template and a game-specific observation prompt (encoding rules and current state), and stores the same core fields for training/analysis (e.g., game/state identifiers, observation text, target action, action label/text/block, optimal-action set, prompt/messages, and target string). The only formatting difference across variants is the thinking flag: reasoning-augmented variants are rendered with thinking enabled, while plain variants are rendered with thinking disabled.

\subsection{SFT Hyperparameters}
\label{app:sft-hyperparameters}

We select Qwen3-8B as the base model for all fine-tuning experiments in this paper based on its strong and consistent zero-shot performance across the four game variants. As shown in Figure~\ref{fig:other_llm_baselines}, Qwen3-8B achieves the highest Elo among the three candidate LLMs in the two open-rule-space games, oTTT and dTTT, while remaining competitive in cTTT and sTTT. At the same time, Qwen3-8B exhibits relatively low trajectory diversity compared with the other LLM baselines, making it a particularly useful starting point for our study: it provides a strong performance baseline while still leaving clear headroom to test whether fine-tuning can recover or increase behavioral diversity. This combination of high competence and comparatively limited diversity makes Qwen3-8B the most suitable base model for analyzing how fine-tuning reshapes both strength and strategic variety across rule variants.

\begin{table}[H]
\centering
\footnotesize
\setlength{\tabcolsep}{6pt}
\begin{tabular}{ll}
\toprule
\textbf{Hyperparameter} & \textbf{Value} \\
\midrule
Base model & Qwen/Qwen3-8B \\
Epochs & 18 (ottt, cttt, dttt) / 12 (sttt) \\
Learning rate & $1\times10^{-4}$ \\
Per-device batch size & 8 \\
Workers (DDP) & 4 \\
Effective batch size & 32 \\
Max sequence length & 1024 \\
Label mode & assistant-only (loss on response tokens) \\
LoRA & $r=32,\alpha=64,\text{dropout}=0.05$ \\
LoRA target modules & q\_proj, k\_proj, v\_proj, o\_proj, gate\_proj, up\_proj, down\_proj \\
Gradient checkpointing & enabled for thinking-trace variants \\
Compute & 4$\times$ NVIDIA H100 GPUs per run \\
\bottomrule
\end{tabular}
\vspace{0.3cm}
\caption{SFT training hyperparameters for Qwen3-8B on the tic-tac-toe family. The epoch budget is reduced for sttt because its larger boards and longer contexts make each training epoch more expensive.}
\label{tab:sft-hyperparameters}
\end{table}

\newpage
\section{Additional Experiments}

\subsection{Other Baseline LLMs}
\label{subsec: app-other-baseline-LLMs}

\begin{figure}[H]
    \centering
    \includegraphics[width=\textwidth]{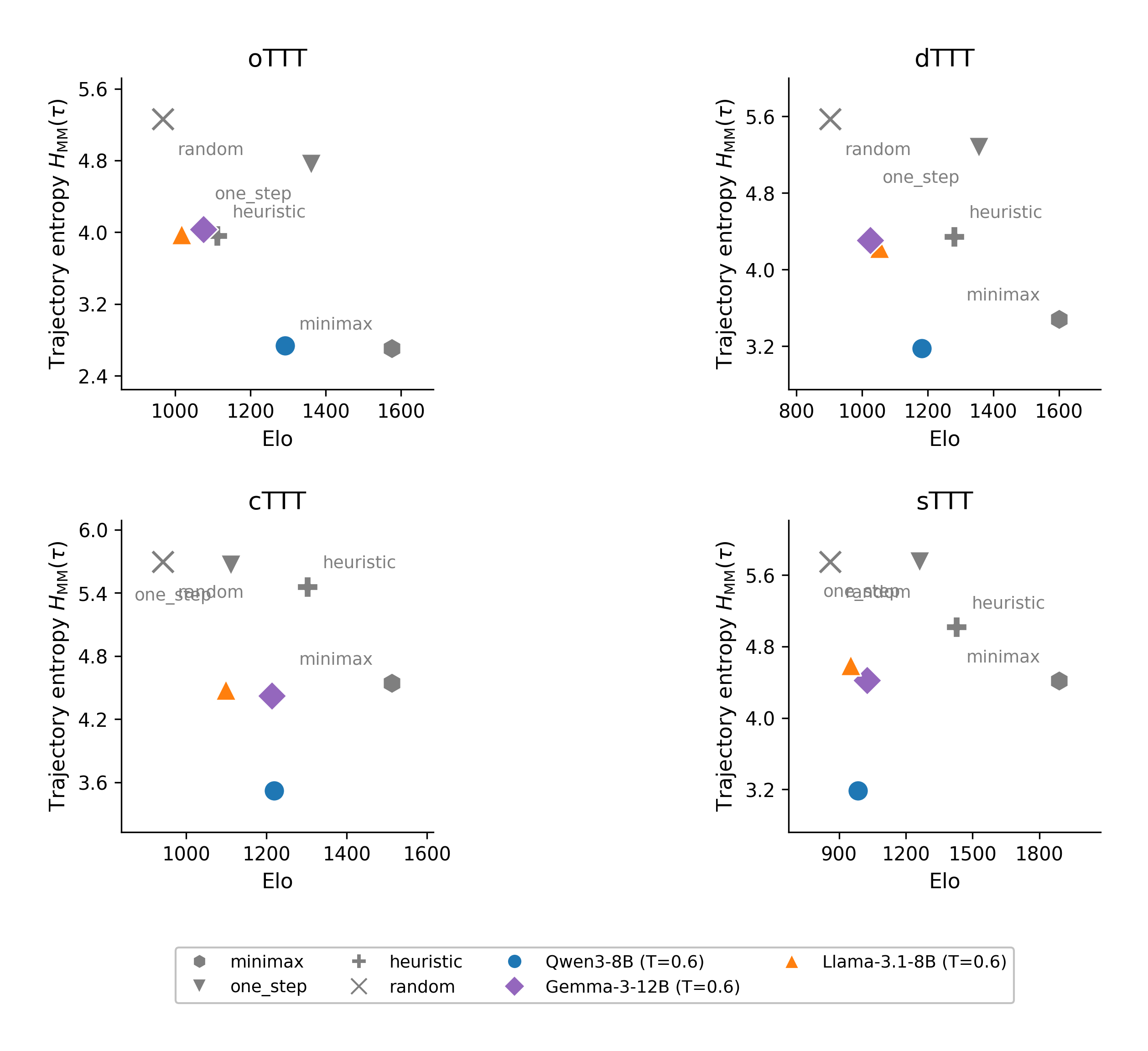}
    \caption{Elo rating versus trajectory entropy $H_{\mathrm{MM}}(\tau)$ for three instruction-tuned LLMs (Qwen3-8B, Gemma-3-12B, Llama-3.1-8B, all at temperature 0.6) compared against deterministic baselines across four game variants.}
    \label{fig:other_llm_baselines}
\end{figure}

We select Qwen3-8B as the base model for all fine-tuning experiments in this paper based on its competitive zero-shot performance across the four game variants. As shown in Figure~\ref{fig:other_llm_baselines}, Qwen3-8B achieves the highest Elo rating among the three candidate LLMs in the two open-rule-space games, oTTT and dTTT. In cTTT and sTTT, Qwen3-8B remains competitive. This consistent strong performance in the games makes Qwen3-8B the most suitable base for studying how fine-tuning shapes game-playing behavior across rule variants.

\subsection{Temperature Study}

A natural first response to trajectory collapse is to raise the temperature of the model at inference time: higher temperature increases sampling randomness and, in principle, should diversify the model's output distribution without requiring any changes to the weights or training data.

\begin{figure}[H]
    \centering
    \includegraphics[width=0.8\textwidth]{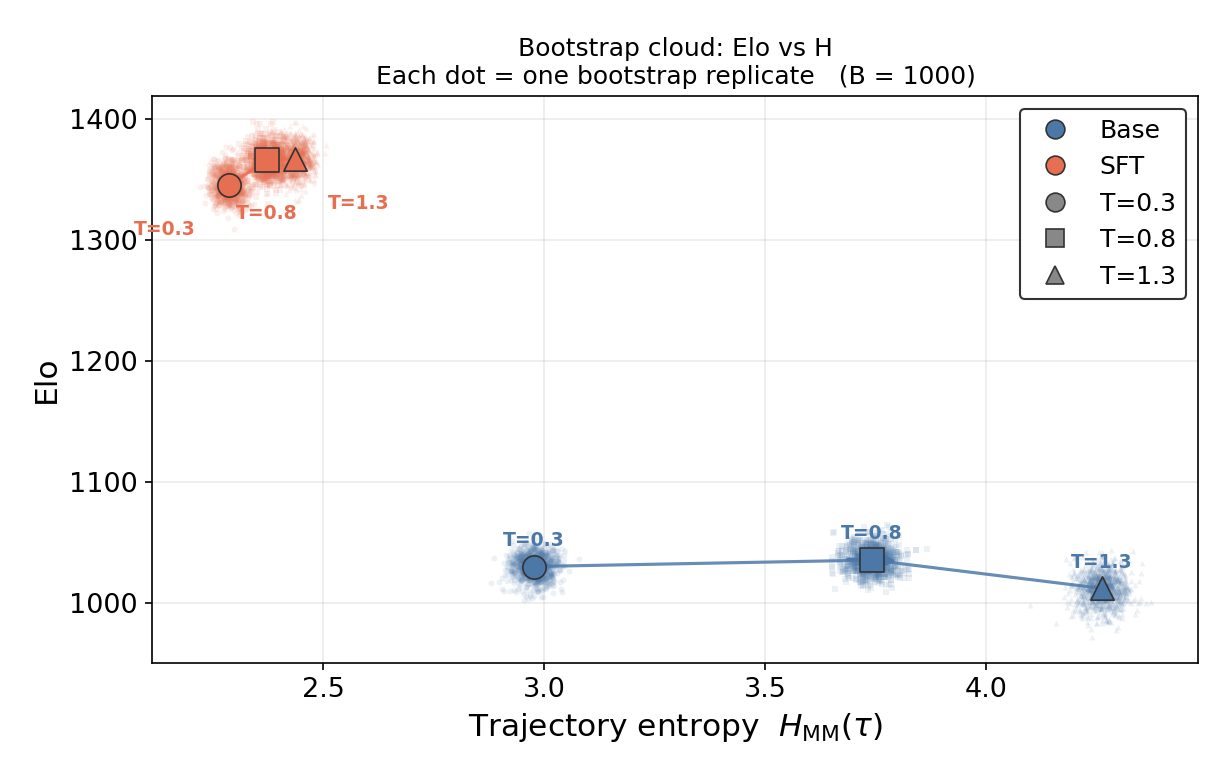}
    \caption{Elo versus Diversity plot for base and plain\_sft Qwen3-8B model}
    \label{fig:games_elo_entropy}
\end{figure}

To test whether temperature alone is sufficient to recover diversity, Figure~\ref{fig:games_elo_entropy} applies a three-temperature sweep $T\in\{0.3,0.8,1.3\}$ to two \texttt{Qwen3-8B} variants - base and plain SFT - using the non-thinking mode in both cases.

\begin{table}[H]
\centering
\footnotesize
\setlength{\tabcolsep}{6pt}
\begin{tabular}{@{}p{0.32\textwidth}c c c c@{}}
\toprule
\textbf{Quantity} & \textbf{Estimate} & \textbf{95\% bootstrap CI} & \textbf{One-sided p} & \textbf{Conclusion} \\
\midrule
$\Delta H_{\mathrm{base}}(0.8-0.3)$
& 0.727 & [0.644, 0.814] & $<0.001$ & Temp increases diversity (Base) \\

$\Delta H_{\mathrm{plain_sft}}(0.8-0.3)$
& 0.088 & [0.023, 0.148] & 0.004 & Temp increases diversity (SFT) \\

$I_{0.8-0.3}=\Delta H_{\mathrm{plain_sft}}-\Delta H_{\mathrm{base}}$
& -0.639 & [-0.744, -0.535] & $<0.001$ & Diminished after SFT \\

\addlinespace
$\Delta H_{\mathrm{base}}(1.3-0.8)$
& 0.481 & [0.392, 0.566] & $<0.001$ & Temp increases diversity (Base) \\

$\Delta H_{\mathrm{sft}}(1.3-0.8)$
& 0.059 & [-0.003, 0.120] & 0.030 & Weak positive gain after SFT \\

$I_{1.3-0.8}=\Delta H_{\mathrm{sft}}-\Delta H_{\mathrm{base}}$
& -0.422 & [-0.531, -0.310] & $<0.001$ & Diminished after SFT \\
\bottomrule
\end{tabular}
\vspace{0.3cm}
\caption{Segment-wise bootstrap tests for temperature effects on diversity. One-sided p-values are empirical bootstrap tail probabilities in the directional hypothesis.}
\label{tab:temp_segment_interaction_bootstrap}
\end{table}

For the base model, temperature has a clear, consistent effect on diversity across both increments: $\Delta H_{\mathrm{base}}(0.8{-}0.3)=0.727$ (95\% CI $[0.644,0.814]$, $p<0.001$) and $\Delta H_{\mathrm{base}}(1.3{-}0.8)=0.481$ (95\% CI $[0.392,0.566]$, $p<0.001$), confirming that each step up in temperature meaningfully widens the base model's action distribution. After supervised fine-tuning, however, the same temperature increments produce dramatically smaller gains: $\Delta H_{\mathrm{sft}}(0.8{-}0.3)=0.088$ ($p=0.004$) and $\Delta H_{\mathrm{sft}}(1.3{-}0.8)=0.059$ ($p=0.030$). The interaction effects $I_{0.8-0.3}=-0.639$ (95\% CI $[-0.744,-0.535]$) and $I_{1.3-0.8}=-0.422$ (95\% CI $[-0.531,-0.310]$) are both strongly negative and significant (Table~\ref{tab:temp_segment_interaction_bootstrap}), confirming that the diversity gain from raising temperature is substantially diminished after SFT across both segments of the sweep. This pattern is also visible in Figure~\ref{fig:games_elo_entropy}: the base model's cloud spans a wide range of the entropy axis as temperature varies, while the SFT model's three conditions cluster tightly, regardless of temperature. Taken together, these results suggest that while temperature scaling can improve diversity on the base model, it does not effectively counteract the distributional compression introduced by SFT---once the model has collapsed, raising temperature alone is insufficient to recover meaningful behavioral diversity.

\subsection{Prompt Engineering: GEPA}
\label{sec: app-prompt-engineering}

To improve the performance of LLMs on playing Tic-Tac-Toe, a natural step is to conduct prompt engineering.
Other literature has explicitly studied how different representations of the board can change a model's performance \citep{topsakal2024evaluating}.

\begin{figure}[H]
    \centering
    \includegraphics[width=0.5\textwidth]{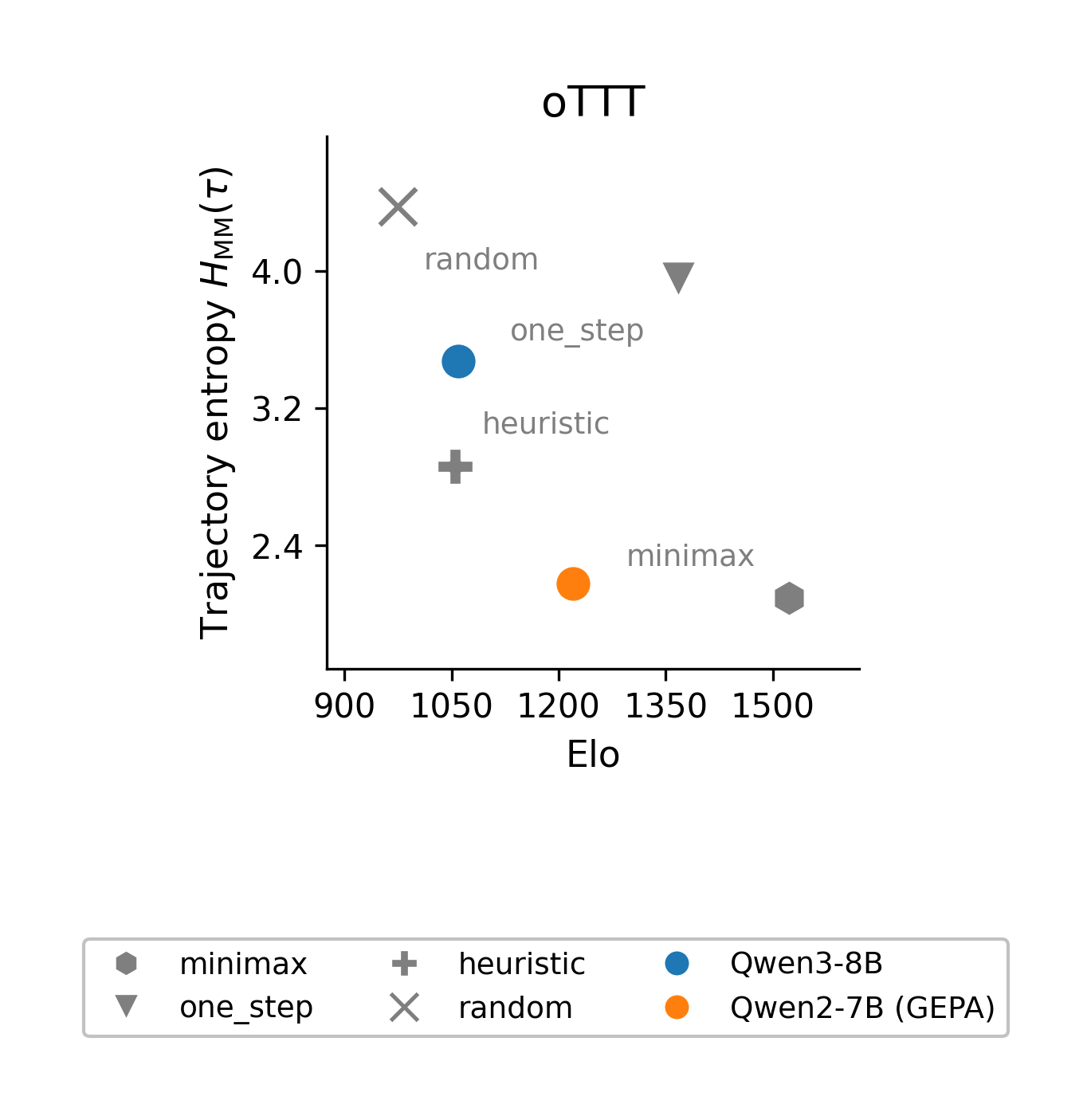}
    \caption{Elo rating versus diversity for \texttt{Qwen3-8B} before and after GEPA prompt optimization.}
    \label{fig:games_gepa}
\end{figure}


GEPA (Genetic-Pareto) is a reflective prompt optimizer introduced by \citet{agrawal2025gepa} that learns high-level behavioral rules from natural language feedback rather than from scalar reward gradients.
Given an AI system and a task metric, GEPA samples rollouts, reflects on them in natural language to diagnose failures and propose prompt revisions, and then tests candidate updates.
To avoid collapsing to a single locally optimal prompt, it maintains a Pareto front of prompts, tracking top performers across different problem instances, and uses genetic-style mutation to combine complementary lessons across generations. The method is reported in \cite{agrawal2025gepa} to require far fewer rollouts than RL-based approaches.

We apply GEPA to the base \texttt{Qwen3-8B} model.
Out of all candidate prompts explored, two produced measurable Elo improvements. The results in Figure~\ref{fig:games_gepa} use Advancement~2, the final and best-performing prompt from the GEPA run.
GEPA moves \texttt{Qwen3-8B} significantly upward in Elo, but the gain comes with a clear reduction in $H_{\mathrm{raw}}$ and $H_{\mathrm{symmetrical}}$: the optimized prompt narrows the model's move distribution, concentrating play on a smaller set of strategies.
$H_{\mathrm{diff}}$ drops further toward zero as well. Against \texttt{heuristic}, the base model produces 96 unique full-game paths out of 200 games (48\%), while under the GEPA prompt that falls to just 6 (3\%). Against \texttt{minimax}, only 2 distinct paths remain out of 200.


Crucially, even with the best prompt GEPA found, the Elo improvement is modest relative to this diversity cost, and the model still falls well short of \texttt{one\_step}.
This suggests that prompt engineering alone is insufficient to close the skill gap, and that improving the model's underlying game knowledge requires direct weight updates.

\end{document}